%% file: arxiv.tex
\documentclass{article} 
\usepackage{iclr2027_conference,times}

\input{math_commands.tex}

\newcommand{\precog}{PreCoG}
\newcommand{\preedit}{PreEdit}
\newcommand{\preqa}{PreQA}

\usepackage{algorithm}
\usepackage{algpseudocode}
\usepackage{amssymb}
\usepackage{graphicx}
\usepackage{booktabs}
\usepackage{xcolor}
\usepackage{colortbl}
\definecolor{comparisontext}{HTML}{666666}
\definecolor{deltagain}{HTML}{E3F1E5}
\definecolor{deltaloss}{HTML}{F8E3E1}
\newcommand{\deltagain}[1]{\begingroup\setlength{\fboxsep}{1pt}\colorbox{deltagain}{\hspace{0.5pt}#1\hspace{0.5pt}}\endgroup}
\newcommand{\deltaloss}[1]{\begingroup\setlength{\fboxsep}{1pt}\colorbox{deltaloss}{\hspace{0.5pt}#1\hspace{0.5pt}}\endgroup}
\usepackage{listings}
\usepackage{placeins}
\usepackage{afterpage}
\usepackage{capt-of}
\usepackage{hyperref}
\usepackage{url}

\fancypagestyle{labpaperfirst}{%
  \fancyhf{}%
  \fancyhead[L]{\includegraphics[width=4.2in]{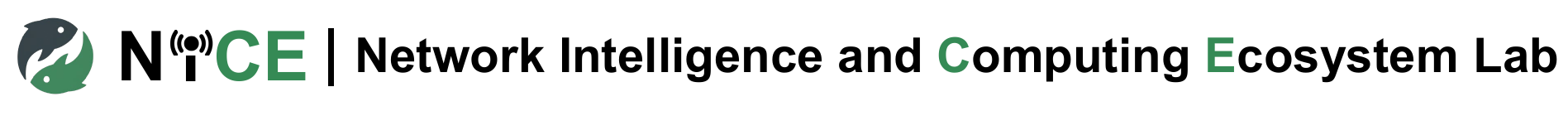}}%
  \fancyfoot[C]{\thepage}
  \renewcommand{\headrulewidth}{0pt}%
}

\fancypagestyle{labpapermain}{%
  \fancyhf{}%
  \fancyhead[L]{arXiv preprint}
  \fancyfoot[C]{\thepage}
  \renewcommand{\headrulewidth}{0pt}%
}

\title{\precog{}: Preview-Edit for Efficient Video \\Correction in Generation-Time
}

\author{
Yilai Liu$^{1}$, Wenqi Pei$^{1}$, Zhifeng Lu$^{2}$,
Shiyuan Zhang$^{1}$, Zongxi Liu$^{1}$, Hongyang Du$^{1}$ \\
$^{1}$NICE Lab, The University of Hong Kong \\
$^{2}$Beijing University of Posts and Telecommunications
}

\iclrfinalcopy
\begin{document}

\maketitle

\pagestyle{labpapermain}
\thispagestyle{labpaperfirst}

\begin{abstract}
Iterative text-to-video generation relies on feedback to correct compositional
and temporal failures. However, existing methods typically require another full
generation run for each correction, repeating computation and potentially disrupting
already correct content. We introduce \textbf{\precog{}}, an iterative framework
for correcting semantic failures during video generation.
\textbf{\preqa{}} diagnoses failures from low-cost previews and constructs
corrective conditions, while \textbf{\preedit{}} edits cached generation trajectories
under these conditions, attenuating orthogonal changes to limit disruption of
already correct content. \precog{} completes generation only for the selected
candidate, reusing its cached trajectory to reduce redundant computation and help
preserve semantic consistency between the preview and the final video.
Experiments show that \precog{} achieves the highest VBench and T2V-CompBench
scores and approaches VSR's leading VS2 score at
$0.56\times$ its inference time. It also outperforms GENMAC on all three quality and
alignment metrics at $0.86\times$ its inference time.
\end{abstract}

\section{Introduction}

Text-to-video generation is rarely a one-shot process in creative workflows.
Users often generate, inspect, revise, and regenerate until one candidate satisfies
the prompt. This loop is costly for two reasons. First, each full rendering requires
expensive denoising steps. Second, even strong models frequently fail on
compositional requirements such as subject presence, action binding, spatial
relations, physical interactions, and temporal progression, often requiring multiple generation attempts before an acceptable result is obtained.
Consequently, substantial computation can be spent on videos that are ultimately rejected.

\afterpage{%
\begin{figure}[t]
    \centering
    \includegraphics[width=\linewidth]{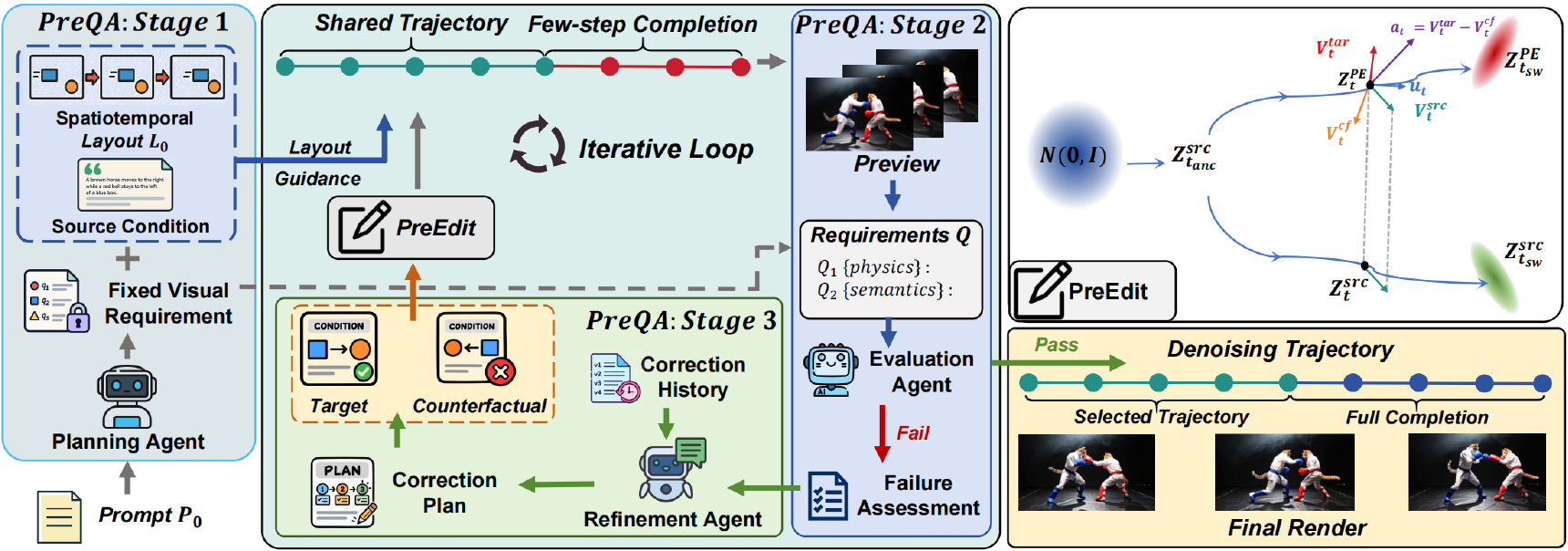}
    \caption{\textbf{Overview of \precog{}.} Preview assessment guides corrective
    edits of a cached source trajectory. Only the selected candidate receives
    full rendering from its cached trajectory.}
    \label{fig:main_overview}
\end{figure}%
}

Recent video-feedback methods make this iterative process more systematic by evaluating
generated videos and using the resulting critique to guide the next attempt.
VQQA~\citep{vqqa} and VISTA~\citep{vista} use generated videos as feedback to
critique and refine prompts for subsequent generation rounds. However, while each
corrected candidate is sampled again under a revised prompt, content that was
already satisfactory in the previous video may drift in the next attempt.
GENMAC~\citep{genmac} further redesigns text prompts, frame-wise layouts, and
guidance scales for compositional generation.
Despite improving how the next attempt is guided, these methods still realize correction through regeneration. Each correction requires another full synthesis run, repeating computation and potentially disrupting already correct content.

These limitations call for a corrective generation process that can intervene early, preserve valid content, and reuse prior computation. Meeting these requirements requires reconsidering when feedback is obtained and how it is applied within the generation process. This leads to
three design principles. First, the system should assess low-cost previews before
full rendering, since early iterations are primarily diagnostic.
Second, feedback should guide corrective edits relative to an existing
source trajectory, avoiding generation from scratch. Since the source may only
partially align with its condition, correction must address unmet requirements
while preserving satisfied ones and limiting unrelated changes.
Third, the computation already
spent on a source video should be treated as a computational asset, with cached denoising states,
source velocities, and solver state supporting subsequent editing and continuation.

Building on these principles, we propose \textbf{\precog{}}, a preview-guided
corrective generation framework for iterative text-to-video
generation. \precog{} consists of two coupled
components. \textbf{\preqa{}} comprises three agents: the Planning Agent
constructs the initial design and fixed visual requirements. The Evaluation
Agent diagnoses unmet requirements in each preview. The Refinement Agent
constructs a matched target--counterfactual condition pair, contrasting the
intended scene with observed failures under shared context. \textbf{\preedit{}} uses their
prediction contrast to steer source-relative updates toward correcting
the diagnosed source--condition misalignments. Blockwise soft projection
emphasizes this correction direction while attenuating orthogonal updates,
aiming to preserve satisfied requirements and limit unrelated changes.
It reuses the source trajectory, velocities, and solver state cached during
generation rather than reconstructing them from a rendered video.
Each correction edits the cached trajectory of the preview with the highest requirement coverage.

The loop alternates between preview assessment and corrective editing until
a candidate satisfies all requirements or the correction budget is exhausted.
The first passing candidate receives full rendering; otherwise, we select the
candidate satisfying the most requirements. This separates low-cost diagnosis from expensive final
rendering, reducing computation spent on unsuccessful iterations.

Our main contributions are summarized as follows:

\begin{itemize}
    \item We propose \textbf{\precog{}}, a preview-guided corrective generation framework that reduces the cost of iterative text-to-video generation. It uses low-cost previews for intermediate correction loops and performs full rendering only for the selected candidate.

    \item We introduce \textbf{\preqa{}}, which coordinates planning, evaluation, and refinement agents to identify unmet visual requirements in low-cost previews and construct matched target and counterfactual conditions for correction.

    \item We propose \textbf{\preedit{}}, a training-free editor that applies blockwise soft projection along a target--counterfactual contrast to cached trajectory updates.

    \item \precog{} achieves the highest VBench and T2V-CompBench scores and approaches VSR's leading VS2 score at $0.56\times$ its inference time, while outperforming GENMAC on all three quality metrics at $0.86\times$ its inference time.
\end{itemize}

\section{Related Work}

\subsection{Video-Feedback Correction for Text-to-Video Generation}

Video-feedback correction evaluates generated videos and converts diagnosed failures into guidance for subsequent refinement.
NeuS-E~\citep{neuse} uses a formal video representation to identify inconsistent events, objects, and frames for targeted editing.
VideoRepair~\citep{videorepair} uses MLLM-generated evaluation questions to detect misalignment, then segments correctly generated entities and selectively regenerates problematic regions. VQQA~\citep{vqqa} dynamically generates visual questions and converts VLM critiques into actionable prompt updates through a black-box language interface.
VISTA~\citep{vista} likewise uses video-level critiques to refine prompts for subsequent generation rounds.
GENMAC~\citep{genmac} extends this loop with multi-agent verification and redesign of prompts, layouts, and guidance scales.
These approaches differ in whether feedback triggers localized editing or another generation round.
\precog{} obtains feedback from low-cost previews and corrects the cached trajectory before full rendering.

\subsection{Inference-Time Search for Video Generation}

Inference-time search improves video generation by exploring multiple candidate
noises, latents, or denoising trajectories during sampling. Diffusion latent beam
search~\citep{latentbeam} ranks intermediate latent candidates using a lookahead estimator of their eventual alignment reward.
Video-T1~\citep{videot1} explores additional noise candidates with verifier feedback; its Tree-of-Frames strategy adaptively expands and prunes video branches autoregressively to reduce search cost.
LatSearch~\citep{latsearch} uses latent reward guidance for more efficient trajectory search. CachedSearch~\citep{cachedsearch} further reduces the
cost of best-of-$N$ video search by using cached rollouts for exploration. These
methods scale quality through wider candidate search, while \precog{} targets the
cost of feedback-based correction.

\subsection{Flow-Based Trajectory Editing}

Flow-based editing studies how to steer generative trajectories between source
and target conditions. FlowEdit introduces an inversion-free and
optimization-free text-based editing method for pretrained flow models by
constructing an ODE that directly maps between source and target prompt
distributions~\citep{flowedit}. Recent video extensions and evaluations adapt
rectified-flow generators for training-free, inversion-free video editing,
including Pyramid-Edit and Wan-Edit in FiVE~\citep{five}.
FlowDirector~\citep{flowdirector} combines direction-aware flow correction, motion-appearance decoupling, and differential averaging to control appearance, temporal consistency, and editing stability.
FlowAnchor~\citep{flowanchor} addresses spatial localization and weakening edit magnitudes through spatial-aware attention refinement and adaptive magnitude modulation.
\preedit{} reuses source states and velocities cached during generation and accounts for source--condition misalignment inherent to correction.
Its target--counterfactual contrast focuses updates on diagnosed failures while attenuating unrelated changes to preserve already correct content.

\begingroup
\begin{figure}[t]
    \centering
    \includegraphics[width=\linewidth]{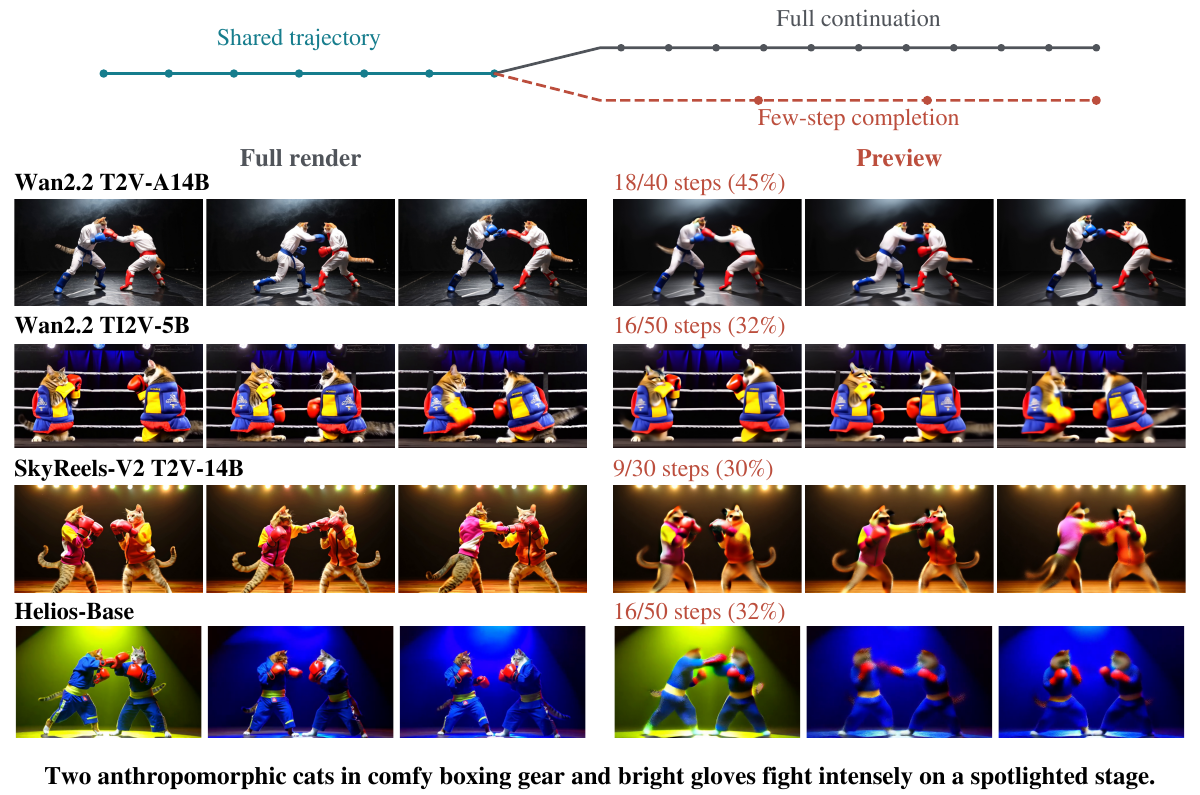}
    \caption{\textbf{Previews and full renders with a shared trajectory.}
    Previews use matched prompts and seeds with few-step completion.
    Percentages denote sampling-step ratios.}
    \label{fig:motivation_observation}
\end{figure}
\endgroup
\input{preview_reliability_arxiv}

\newcommand{\arxivexperimentfloats}{%
    \input{main_results_arxiv}    \input{qualitative_sample03_04_arxiv}    \input{ablations_arxiv}    \input{ablation_qualitative_arxiv}}
\section{Methodology}
\input{Methodology_arxiv}

%
%

\section{Experiments}
\label{sec:experiments}
\input{Experiments}

\section{Conclusion}
\label{sec:conclusion}
\input{Conclusion}

\subsection*{AI use statement}
We used generative AI tools to polish the language and improve the clarity
and readability of the manuscript. The authors take responsibility for the
final content of the paper.

\subsection*{Ethics statement}
All four participants were informed of the purpose of the human evaluation
beforehand and voluntarily agreed to participate.

\bibliography{iclr2027_conference}
\bibliographystyle{iclr2027_conference}

\clearpage
\appendix
\input{Appendix}

\end{document}

%% file: math_commands.tex
\usepackage{amsmath,amsfonts,bm}

\def\eqref#1{equation~\ref{#1}}

\def\1{\bm{1}}

\DeclareMathAlphabet{\mathsfit}{\encodingdefault}{\sfdefault}{m}{sl}
\SetMathAlphabet{\mathsfit}{bold}{\encodingdefault}{\sfdefault}{bx}{n}



%% file: preview_reliability_arxiv.tex
\begin{figure}[t]
    \centering
    \includegraphics[width=0.94\linewidth]{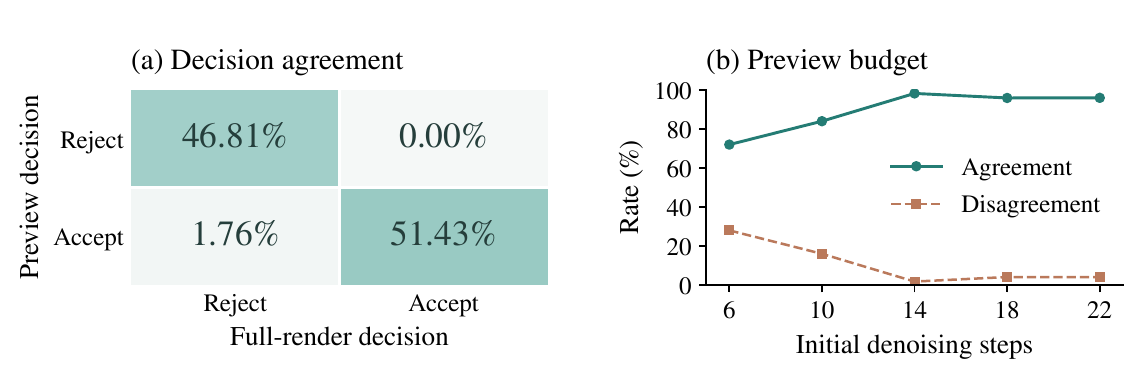}
    \caption{\textbf{Preview reliability.} Left: preview/full decision
    agreement using Gemini 3.1 Pro~\citep{gemini31pro} in separate requests. Cells show percentages of all preview/full pairs.
    Right: agreement and disagreement versus the number of initial denoising steps,
    with two bridge and two clean steps.
    Disagreement is the fraction of pairs with different preview and full-render decisions.}
    \label{fig:preview_reliability}
\end{figure}

%% file: main_results_arxiv.tex
\begin{table}[t]
\centering

\setlength{\tabcolsep}{2pt}
\caption{\textbf{Quantitative comparison.} We report VBench, T2V-CompBench (T2V-Comp.), VS2, and inference time. $\Delta$: change from direct generation. \textbf{Bold}: best. \underline{Underlined}: second best.}
\label{tab:main_results}
\begin{tabular*}{\linewidth}{@{\extracolsep{\fill}}lr@{\extracolsep{0pt}\hspace{4pt}}>{\color{comparisontext}}r@{\extracolsep{\fill}\hspace{10pt}}r@{\extracolsep{0pt}\hspace{4pt}}>{\color{comparisontext}}r@{\extracolsep{\fill}\hspace{10pt}}r@{\extracolsep{0pt}\hspace{4pt}}>{\color{comparisontext}}r@{\extracolsep{\fill}\hspace{10pt}}r@{\extracolsep{0pt}\hspace{4pt}}>{\color{comparisontext}}r@{}}
\toprule
Method & \multicolumn{2}{c}{VBench (\% $\uparrow$)} & \multicolumn{2}{c}{T2V-Comp. (\% $\uparrow$)} & \multicolumn{2}{c}{VS2 ($\uparrow$)} & \multicolumn{2}{c}{Time (s $\downarrow$)} \\
\cmidrule(lr){2-3}\cmidrule(lr){4-5}\cmidrule(lr){6-7}\cmidrule(lr){8-9}
 & Value & $\Delta$ & Value & $\Delta$ & Value & $\Delta$ & Value & Ratio \\
\midrule
Direct generation & 82.06 & 0.00 & 55.27 & 0.00 & 3.708 & 0.000 & \textbf{895.0} & 1.00$\times$ \\
\midrule
\multicolumn{9}{l}{\textit{Inference-time search}} \\
Full best-of-$N$ ($N=4$) & 81.29 & \deltaloss{-0.77} & 52.62 & \deltaloss{-2.65} & 3.751 & \deltagain{+0.043} & 3580.0 & \deltaloss{4.00$\times$} \\
CachedSearch ($N=8$) & 82.11 & \deltagain{+0.05} & 55.02 & \deltaloss{-0.25} & 3.803 & \deltagain{+0.095} & 4338.9 & \deltaloss{4.85$\times$} \\
\midrule
\multicolumn{9}{l}{\textit{Agentic multi-loop generation}} \\
VQQA & 80.20 & \deltaloss{-1.86} & 51.94 & \deltaloss{-3.33} & 3.718 & \deltagain{+0.010} & 3166.8 & \deltaloss{3.54$\times$} \\
VSR & 80.43 & \deltaloss{-1.63} & \underline{61.82} & \deltagain{+6.55} & \textbf{3.973} & \deltagain{+0.265} & 3620.3 & \deltaloss{4.05$\times$} \\
GENMAC & \underline{82.76} & \deltagain{+0.70} & 56.76 & \deltagain{+1.49} & 3.659 & \deltaloss{-0.049} & 2350.5 & \deltaloss{2.63$\times$} \\
\midrule
\textbf{\precog{} (ours)} & \textbf{83.49} & \deltagain{+1.43} & \textbf{62.28} & \deltagain{+7.01} & \underline{3.854} & \deltagain{+0.146} & \underline{2022.9} & \deltaloss{2.26$\times$} \\
\bottomrule
\end{tabular*}
\end{table}

%% file: qualitative_sample03_04_arxiv.tex
\begin{figure}[t]
    \normalsize
    \centering
    \includegraphics[width=\linewidth]{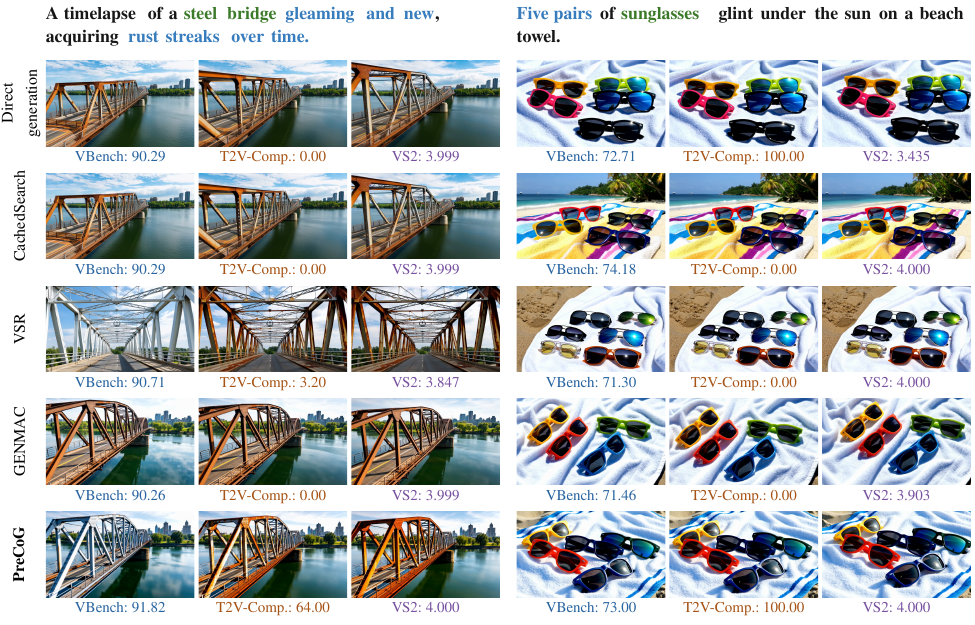}
    \caption{\textbf{Qualitative comparison on compositional prompts.}
    Generated videos depict temporal attribute changes (left) and object
    counting (right), with three uniformly sampled frames per video.
    Colored labels report VBench, T2V-CompBench (T2V-Comp.) for the corresponding category, and
    VS2 alignment scores for the complete videos.}
    \label{fig:qualitative_sample03_04}
\end{figure}

%% file: ablations_arxiv.tex
\begin{table}[t]
\centering

\setlength{\tabcolsep}{2pt}
\caption{\textbf{Component ablations of \precog{}.} Metrics follow Table~\ref{tab:main_results}. VS2 uses soft alignment. $\Delta$ denotes change from direct generation. \textbf{Bold}: best. \underline{Underlined}: second best.}
\label{tab:ablations}
\begin{tabular*}{\linewidth}{@{\extracolsep{\fill}}lr@{\extracolsep{0pt}\hspace{4pt}}>{\color{comparisontext}}r@{\extracolsep{\fill}\hspace{5pt}}r@{\extracolsep{0pt}\hspace{4pt}}>{\color{comparisontext}}r@{\extracolsep{\fill}\hspace{5pt}}r@{\extracolsep{0pt}\hspace{4pt}}>{\color{comparisontext}}r@{\extracolsep{\fill}\hspace{5pt}}r@{\extracolsep{0pt}\hspace{4pt}}>{\color{comparisontext}}r@{}}
\toprule
Variant & \multicolumn{2}{c}{VBench (\% $\uparrow$)} & \multicolumn{2}{c}{T2V-Comp. (\% $\uparrow$)} & \multicolumn{2}{c}{VS2 ($\uparrow$)} & \multicolumn{2}{c}{Time (s $\downarrow$)} \\
\cmidrule(lr){2-3}\cmidrule(lr){4-5}\cmidrule(lr){6-7}\cmidrule(lr){8-9}
 & Value & $\Delta$ & Value & $\Delta$ & Value & $\Delta$ & Value & Ratio \\
\midrule
Direct generation & 82.06 & 0.00 & 55.27 & 0.00 & 3.708 & 0.000 & \textbf{895.0} & 1.00$\times$ \\
\midrule
\multicolumn{9}{c}{\textit{\preqa{}}} \\
Evaluation w/ full video & \textbf{83.56} & \deltagain{+1.50} & \underline{60.66} & \deltagain{+5.39} & \textbf{3.861} & \deltagain{+0.153} & 2809.8 & \deltaloss{3.14$\times$} \\
Planning w/o planning & 82.79 & \deltagain{+0.73} & 56.50 & \deltagain{+1.23} & 3.644 & \deltaloss{-0.064} & \underline{1752.7} & \deltaloss{1.96$\times$} \\
Refinement w/o target refinement & 82.48 & \deltagain{+0.42} & 53.76 & \deltaloss{-1.51} & 3.486 & \deltaloss{-0.222} & 1986.4 & \deltaloss{2.22$\times$} \\
\midrule
\multicolumn{9}{c}{\textit{\preedit{}}} \\
Regeneration instead of editing & 82.90 & \deltagain{+0.84} & 56.10 & \deltagain{+0.83} & 3.735 & \deltagain{+0.027} & 2354.1 & \deltaloss{2.63$\times$} \\
w/o counterfactual projection & 83.20 & \deltagain{+1.14} & 56.80 & \deltagain{+1.53} & 3.760 & \deltagain{+0.052} & 1973.3 & \deltaloss{2.20$\times$} \\
\midrule
\textbf{\precog{} (full)} & \underline{83.49} & \deltagain{+1.43} & \textbf{62.28} & \deltagain{+7.01} & \underline{3.854} & \deltagain{+0.146} & 2022.9 & \deltaloss{2.26$\times$} \\
\bottomrule
\end{tabular*}
\end{table}

%% file: ablation_qualitative_arxiv.tex
\begin{figure}[t]
    \normalsize
    \centering
    \includegraphics[width=\linewidth]{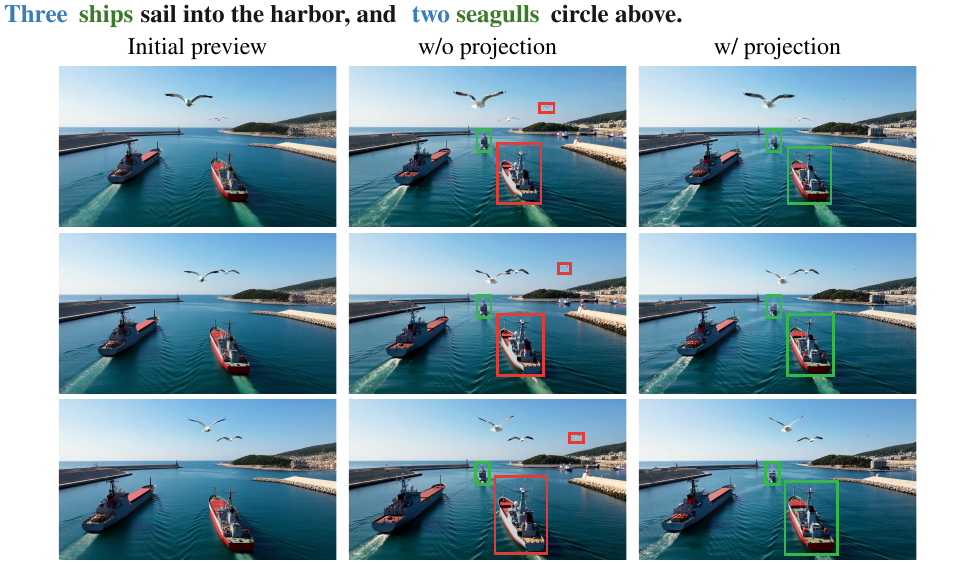}
    \caption{\textbf{Counterfactual projection ablation.}
    Initial preview, w/o counterfactual projection, and w/ counterfactual
    projection, from left to right.
    Rows show the first, middle, and last frames.}
    \label{fig:ablation_qualitative}
\end{figure}

%% file: Methodology_arxiv.tex
\input{Motivation}

\subsection{\precog{}}
\label{sec:method_overview}


Given a prompt $P_0$, \precog{} alternates semantic assessment of low-cost
previews with corrective trajectory editing within a fixed budget
(Figure~\ref{fig:main_overview}).
Because the source trajectory may only partially align with its condition,
correction targets unmet requirements while aiming to preserve satisfied ones.
\textbf{\preqa{}} (Section~\ref{sec:precogqa}) constructs the initial design and forms matched target
and counterfactual conditions from preview evidence. \textbf{\preedit{}} (Section~\ref{sec:precogedit})
uses their prediction contrast to modulate source-relative updates through
blockwise soft projection. Each correction edits the cached trajectory of the
preview with the highest requirement coverage.
Only the selected candidate receives full rendering, reusing its edited
trajectory segment when available. Algorithm~\ref{alg:precogqa} summarizes the framework.
The following components detail the three agents, editing, and selection.

\subsubsection{\preqa{}: Planning, Evaluation, and Refinement Agents}
\label{sec:precogqa}

\preqa{} follows three stages, each handled by a corresponding agent:
planning, evaluation, and refinement. The Planning Agent runs once.
The Evaluation and Refinement Agents then alternate with \preedit{}
in the preview correction loop.

\paragraph{Stage 1: Planning Agent.}
We reuse GENMAC's~\citep{genmac} initial generation planning to map the original
prompt $P_0$ to a source condition $P_{\mathrm{src}}$ and a spatiotemporal
layout $\mathcal{L}_0$, specifying object identities, locations, and motion.
The layout guides generation of the initial preview, whose trajectory is
cached for subsequent correction.

In addition, we decompose $P_0$ into a fixed set of atomic visual requirements
$\mathcal{Q}=\{q_j\}_{j=1}^{M}$, covering semantic facts, spatial relations,
and physical or temporal events. Constructed directly from the original prompt,
these requirements assess generated content rather than the planning output.
Each requirement retains subject and reference entities to preserve bindings.
The set stays fixed across rounds so prompt refinement cannot relax the
evaluation target.

\paragraph{Stage 2: Evaluation Agent.}
The Evaluation Agent assesses preview $V_i$ against each requirement in $\mathcal{Q}$,
recording its satisfaction verdict and the corresponding observed fact
in $\mathcal{R}_i$. Here $i=0$ denotes the initial preview, and $i>0$
indexes corrective candidates. With binary verdicts $a_{i,j}\in\{0,1\}$,
we define the unmet requirement set $\mathcal{F}_i$ and requirement coverage
$S_i$, the number of satisfied requirements:
\begin{equation}
\mathcal{F}_i=\{q_j\in\mathcal{Q}:a_{i,j}=0\},
\qquad S_i=\sum_{j=1}^{M}a_{i,j},
\qquad \operatorname{Accept}(V_i)\iff\mathcal{F}_i=\emptyset.
\label{eq:preview_acceptance}
\end{equation}
A passing preview is selected for full rendering. Otherwise, we choose the
preview with the highest requirement coverage,
$b_i=\arg\max_{0\leq j\leq i}S_j$, breaking ties in favor of earlier iterations.
The next correction uses the assessment $\mathcal{R}_{b_i}$ and cached trajectory
of $V_{b_i}$, subject to the remaining budget (Section~\ref{sec:preview_render}).

\paragraph{Stage 3: Refinement Agent.}
Using the assessment $\mathcal{R}_{b_i}$ and correction history $\mathcal{H}_i$,
the agent forms a joint repair plan $\mathcal{E}_i$ for failed requirements across
all affected subjects. Its correction description $L_i$ is combined with the
remaining requirements to form the target condition $P_i^{\mathrm{tar}}$.
The generation conditions are $P_0^{\mathrm{gen}}=P_{\mathrm{src}}$ and
$P_{i+1}^{\mathrm{gen}}=P_i^{\mathrm{tar}}$.

The counterfactual condition $P_i^{\mathrm{cf}}$ replaces only the selected
attributes in $P_i^{\mathrm{tar}}$ with their observed failure values, keeping
entity identities and other scene context fixed. After validation,
\textbf{\preedit{}} uses the difference between predictions under the two
conditions to guide correction. Since the conditions differ only in the diagnosed
attributes, this contrast estimates a direction from the observed failure toward
the desired state. The resulting candidate returns to Stage~2 for assessment.

%
    \input{main_results_arxiv}    \input{qualitative_sample03_04_arxiv}    \input{ablations_arxiv}    \input{ablation_qualitative_arxiv}
\subsubsection{\preedit{}: Cached Trajectory Editing with Soft Projection}
\label{sec:precogedit}

\input{PreCoGEdit_editing}

\subsubsection{Preview Selection and Full Rendering}
\label{sec:preview_render}

\paragraph{Shared trajectory, different completion budgets.}
Preview and full rendering share the initial trajectory through $K_p$ steps,
or the corrected trajectory through $k_b$, before following different completion schedules.
For the selected candidate $i^*$, we restore its cached latent and solver state
$\mathcal B_{i^*}$ and complete the native schedule under $P_{i^*}^{\mathrm{gen}}$.
This uses the source condition for the initial candidate and the refined target
condition for an edited candidate. Reusing the shared trajectory avoids repeated
computation and helps preserve semantic consistency between preview and full video.

\paragraph{Stopping and candidate selection.}
We select the first preview satisfying all requirements in Eq.~\ref{eq:preview_acceptance}.
If none passes within the budget of $N$ previews, or valid correction
conditions cannot be obtained, we select the preview with the highest requirement coverage $S_i$,
breaking ties in favor of earlier iterations. Only the selected candidate
receives full rendering.

%% file: Motivation.tex
\subsection{Motivation}
\label{sec:motivation}

Existing approaches to anticipating the outcome of full generation at low
cost include preview and lookahead methods that use few-step solvers to
produce approximate outputs~\citep{consistencysolver,latentbeam}.
Such approximations can inform candidate evaluation. However, these previews
do not inherently preserve the native denoising trajectory or its solver state,
limiting both computational reuse and subsequent trajectory-based interventions
to guide generation. This raises a question: \emph{can an informative
preview share an initial trajectory segment with full rendering?}

\paragraph{Accelerating denoising completion.}
Diffusion models recover coarse structure before fine detail~\citep{masf},
suggesting that later, lower-noise steps can be reduced for preview assessment.
We retain the first $K_p$ updates of the native scheduler, such as
UniPC~\citep{unipc}, and reduce subsequent denoising steps. We compare full rendering with two-step completion
after the shared trajectory segment, using matched conditioning and seeds.

Figure~\ref{fig:motivation_observation} shows selected text-conditioned
examples from Wan2.2-T2V-A14B, Wan2.2-TI2V-5B~\citep{wan},
SkyReels-V2-T2V-14B~\citep{skyreelsv2}, and Helios-Base~\citep{helios}. The shortened runs retain
recognizable subjects, scene arrangement, and broadly corresponding
action stages, while fine detail and some poses differ. These examples suggest that the semantic structure governing prompt
alignment is largely established before full denoising is complete. Additional image-conditioned examples are provided in
Appendix~\ref{app:motivation_probe}.

\paragraph{Implication for \precog{}.}
This observation motivates using low-cost previews to guide correction
before full rendering. \precog{} combines preview-based semantic assessment
with corrective trajectory editing, reusing the selected candidate's
trajectory segment to complete the final video.

%% file: PreCoGEdit_editing.tex
Video correction starts from a generated trajectory that may only partially
satisfy its prompt. It must repair unmet requirements while preserving
satisfied content and limiting unrelated changes. We contrast target and
counterfactual conditions under shared scene context to obtain a correction
direction, then use soft projection to modulate updates along this direction.

\paragraph{Source reference from the selected candidate.}
At round $i$, we reuse the cache $\mathcal{C}_{b_i}$ of the preview with the
highest requirement coverage (Section~\ref{sec:precogqa}). It stores the actual latent states
$\mathbf z_k^{\mathrm{src}}$, applied velocities $\mathbf v_k^{\mathrm{src}}$,
and solver state from generation or editing. This avoids trajectory
reconstruction and repeated source-model evaluations.
During the editing window $W_{\mathrm{edit}}=[k_a,k_b)$, we express the edited
state as $\mathbf z_{i,k}^{\mathrm{edit}}=\mathbf z_k^{\mathrm{src}}+
\boldsymbol\delta_{i,k}$, with $\boldsymbol\delta_{i,k_a}=\mathbf0$.
The branch follows the source before $k_a$, and $k_b\leq K_p$ keeps editing
within the cached trajectory.

\paragraph{Counterfactual-guided soft projection.}
At the edited state and timestep $t_k$, we evaluate the target and
counterfactual conditions with the same CFG scale $g_{\mathrm{tar}}$,
obtaining $\mathbf v_{i,k}^{\mathrm{tar}}$ and $\mathbf v_{i,k}^{\mathrm{cf}}$.
The regular source-relative update and contrastive direction are
\begin{equation}
\mathbf d_{i,k}=\mathbf v_{i,k}^{\mathrm{tar}}-\mathbf v_k^{\mathrm{src}},
\qquad
\mathbf a_{i,k}=\mathbf v_{i,k}^{\mathrm{tar}}-\mathbf v_{i,k}^{\mathrm{cf}}.
\label{eq:source_relative_flow}
\end{equation}
The shared unconditional CFG response cancels in $\mathbf a_{i,k}$.
We apply projection independently to disjoint spatiotemporal blocks $B$,
including all channels within each block:
\begin{equation}
\alpha_{i,k,B}=\operatorname{clip}\!\left(
\frac{\langle\mathbf d_{i,k,B},\mathbf a_{i,k,B}\rangle}
{\|\mathbf a_{i,k,B}\|_2^2+\epsilon},0,\alpha_{\max}\right),
\qquad
\mathbf p_{i,k,B}=\alpha_{i,k,B}\mathbf a_{i,k,B}.
\label{eq:counterfactual_projection}
\end{equation}
The nonnegative coefficient preserves the contrastive direction, while
$\alpha_{\max}$ limits the projected magnitude.
Within the early window $W_{\mathrm{cf}}=[k_a,k_c)\subseteq W_{\mathrm{edit}}$,
we blend the assembled projection $\mathbf p_{i,k}$ with the regular update:
\begin{equation}
\mathbf u_{i,k}=(1-\beta)\mathbf p_{i,k}+\beta\mathbf d_{i,k},
\qquad \beta\in[0,1].
\label{eq:soft_counterfactual_update}
\end{equation}
Here $\beta=0$ gives strict projection and $\beta=1$ gives regular editing.
Ignoring $\epsilon$, this blend cannot increase squared distance to any
correction in the bounded segment
$\{\lambda\mathbf a:0\leq\lambda\leq\alpha_{\max}\}$ for each block
(Appendix~\ref{app:projection_geometry}). Its semantic benefit therefore
depends on whether the contrast captures the required correction.
Outside $W_{\mathrm{cf}}$, we use $\mathbf u_{i,k}=\mathbf d_{i,k}$
without the extra counterfactual prediction.

\paragraph{Displacement integration and continuation.}
With edit strength $\gamma$ and decreasing noise levels $\sigma_k$, we
integrate the displacement using Euler and attach it to the next cached state:
\begin{equation}
\begin{aligned}
\boldsymbol\delta_{i,k+1}
&=\boldsymbol\delta_{i,k}+(\sigma_{k+1}-\sigma_k)\gamma\mathbf u_{i,k},\\
\mathbf z_{i,k+1}^{\mathrm{edit}}
&=\mathbf z_{k+1}^{\mathrm{src}}+\boldsymbol\delta_{i,k+1},
\qquad k\in W_{\mathrm{edit}}.
\end{aligned}
\label{eq:edited_solver_step}
\end{equation}
Setting $\gamma=0$ recovers the source trajectory.
At $k_b$, we save the edited latent and solver state as $\mathcal B_{i+1}$
and continue denoising under the same refined target condition.
Section~\ref{sec:preview_render} details continuation.

%% file: Experiments.tex
\newcommand{\mockresults}{\textit{Illustrative data only. Not experimental results.}}

\subsection{Experimental Settings}
\label{sec:experimental_settings}

\paragraph{Implementation details.}
All methods use Wan2.2-T2V-A14B~\citep{wan} with a native $40$-step schedule to generate
81-frame videos at $832\times480$ resolution. Methods requiring a VLM use
Qwen3.8-27B~\citep{qwen38} for scoring, planning, and feedback.
\precog{} uses a $14+2+2$ preview schedule and allows at most four editing rounds.
Appendix~\ref{app:experimental_details} provides detailed schedules, editing
hyperparameters, and baseline configurations.

\paragraph{Benchmarks and Evaluation Metrics.}
We evaluate on T2V-CompBench prompts~\citep{t2vcompbench}, measuring visual
quality with a six-dimension VBench~\citep{vbench} mean,
prompt alignment with T2V-CompBench and VideoScore2 (VS2)~\citep{videoscore2},
and efficiency with end-to-end inference time.
Appendix~\ref{app:metric_breakdown} defines the metrics.
Human evaluation (Appendix~\ref{sec:exp_human}) shows that \precog{} outperforms
all three feedback baselines in visual quality and prompt alignment, although
most comparisons with VSR are judged as ties.

\input{Experiments_Preview_Reliability}
\input{Experiments_Main}
\input{Experiments_Ablation}

%% file: Experiments_Preview_Reliability.tex
\subsection{Preview Reliability}
\label{sec:preview_reliability}
We select fully rendered videos from the initial generation round and pair
each with its corresponding preview. For this diagnostic only,
Gemini 3.1 Pro~\citep{gemini31pro} serves as an independent evaluator, assessing
each video in a separate, stateless request under the same fixed requirements, without access
to its paired result or preview/full identity.
Figure~\ref{fig:preview_reliability}(a) shows 98.24\% decision agreement.
The remaining 1.76\% comprises accepted previews with rejected full continuations.
This comparison covers initial candidates under a fixed generation seed.

Figure~\ref{fig:preview_reliability}(b) examines sensitivity to the shared
trajectory length, with two bridge and two clean steps for preview completion.
Agreement rises from 72.0\% at 6 native steps to 98.24\% at 14 steps,
while disagreement falls from 28.0\% to 1.76\%.
These results support 14 native steps as a balance between assessment
consistency and preview cost in this comparison.

%% file: Experiments_Main.tex
\subsection{Main Result}
\label{sec:exp_main}

\paragraph{Quantitative Results.}
Table~\ref{tab:main_results} compares visual quality, prompt alignment, and
inference time. \precog{} improves all three
quality and alignment metrics over direct generation, gaining 1.43 points in
VBench, 7.01 points in T2V-Comp., and 0.146 in VS2. Among search and feedback methods, \precog{} is fastest,
requiring $2.26\times$ the time of direct generation.

\precog{} achieves the highest VBench and T2V-Comp. scores, exceeding VSR by
3.06 and 0.46 points, respectively. VSR leads in VS2 by 0.119, but \precog{}
requires only $0.56\times$ its inference time, offering competitive alignment
with higher visual quality at substantially lower cost.
Compared with GENMAC, \precog{} improves all three metrics while reducing
reported inference time by 13.9\%.
Figure~\ref{fig:quality_efficiency_tradeoff} in Appendix~\ref{app:metric_breakdown}
visualizes these quality--cost trade-offs.
Appendix~\ref{app:metric_breakdown} further shows leading scores in motion
binding, action binding, and generative numeracy.

Greater search expenditure does not consistently improve compositional
alignment. Full best-of-$N$ and CachedSearch require
$4.00\times$ and $4.85\times$ the time of direct generation, respectively,
but their T2V-Comp. scores are lower by 2.65 and 0.25 points, despite gains
in VS2. This motivates failure-guided correction rather than candidate selection alone.

\paragraph{Qualitative Results.}
Figure~\ref{fig:qualitative_sample03_04} compares \precog{} with the baselines
on two compositional prompts. In the left example, direct generation,
CachedSearch, and GENMAC show rust on the bridge from the first frame,
missing the requested transition from ``gleaming and new'' to rust-streaked.
\precog{} depicts a clean, silvery bridge that develops a rust-colored appearance,
with consistent bridge structure and background layout across the sampled frames.

In the right example, direct generation already shows the requested five pairs
of sunglasses, whereas VSR and GENMAC produce six and four pairs, respectively.
This comparison illustrates how iterative generation can lose content that was
already correct. \precog{} retains the correct count of five with the requested
sunlight glint. Together, these examples illustrate the importance of repairing
unmet requirements while retaining satisfied ones.





%% file: Experiments_Ablation.tex
\subsection{Component Ablations}
\label{sec:exp_ablation}

Ablations in Table~\ref{tab:ablations} use matched prompts, source seeds, and
correction budgets. Initial designs and source sampling are also matched except
when ablating planning.

\paragraph{\preqa{}.}
Assessing full videos increases VBench and VS2 by only 0.07 and 0.007,
respectively, while reducing T2V-Comp. by 1.62 points and raising runtime
from 2022.9 to 2809.8 seconds. Preview assessment therefore reduces runtime
by 28.0\% with comparable VBench and VS2 and higher compositional alignment.
Removing planning reduces T2V-Comp. from 62.28 to 56.50 and VS2 from 3.854 to
3.644, supporting its role in compositional alignment. Without target refinement,
these scores fall further to 53.76 and 3.486, below direct generation.
This variant retains diagnosis and failure-condition construction but uses the
original prompt as the target. Failure diagnosis must therefore translate into
an explicit repair condition.

\paragraph{Correction mechanism.}
Regeneration replaces editing with full generation under the same target
condition, correction budget, and selection rule. Removing projection retains
cached editing but uses unprojected updates. In Figure~\ref{fig:ablation_qualitative},
both editing variants introduce the missing third ship. Without projection,
the foreground ship changes appearance and an additional seagull appears.
Counterfactual projection retains the count correction while better preserving
unrelated content.

Additional results in Appendix~\ref{app:additional_ablation} further examine
the generality of \precog{} across feedback models, generation backbones,
and conditioning tasks. With sufficiently capable VLMs, \precog{} consistently
improves visual quality and alignment over direct generation, indicating that
its effectiveness is not tied to a particular feedback model. Similar gains
are observed across four T2V and two I2V backbone--task configurations,
suggesting that the proposed preview-guided correction mechanism transfers
across different generation settings.

%% file: Conclusion.tex
We presented \precog{}, a framework for correcting semantic failures during
iterative video generation. Low-cost previews support diagnosis without
completing every candidate, while cached trajectories provide reusable starting
points for correction and final rendering. Experiments show a favorable balance
of visual quality, prompt alignment, and inference cost. Qualitative comparisons
further illustrate how targeted editing can repair unmet requirements while
limiting disruption of already correct content. Gains across multiple feedback
models, backbones, and T2V/I2V settings further support the applicability of this
approach. This correction process requires no additional training of the video
generator. \precog{} makes correction part of the generation process: previews
guide targeted edits to cached trajectories, and full rendering is reserved for
the selected candidate.

%% file: Appendix.tex
\section{Motivation Probe Settings}
\label{app:motivation_probe}

Figures~\ref{fig:motivation_observation} and~\ref{fig:motivation_observation_i2v}
compare full renders and previews using matched prompts, seeds, and output specifications within each pair.
The probes cover Wan2.2, SkyReels-V2~\citep{skyreelsv2}, and Helios-Base~\citep{helios}; I2V pairs additionally share the reference image.
Each preview retains an initial segment of the native trajectory and uses few-step completion.

\paragraph{Text-to-video probes.}
The T2V probes use seed 42. Figure~\ref{fig:motivation_observation} reports
preview/full budgets of $18/40$ (45\%), $16/50$ (32\%), $9/30$ (30\%),
and $16/50$ (32\%) for Wan2.2-T2V-A14B, Wan2.2-TI2V-5B,
SkyReels-V2-T2V-14B, and Helios-Base, respectively.
Displayed frames are sampled at 25\%, 75\%, and 100\% of each clip's frame-index range.

\paragraph{Image-to-video probes.}
For Wan2.2-I2V-A14B, Wan2.2-TI2V-5B, SkyReels-V2-I2V-14B, and Helios-Base,
Figure~\ref{fig:motivation_observation_i2v} reports preview/full budgets of
$19/50$ (38\%), $16/50$ (32\%), $9/30$ (30\%), and $16/50$ (32\%), respectively.
Displayed frames have zero-based indices 20, 60, and 80.
Helios step counts are per segment in both probes.

\begingroup
\begin{figure}[htbp]
    \centering
    \includegraphics[width=\linewidth]{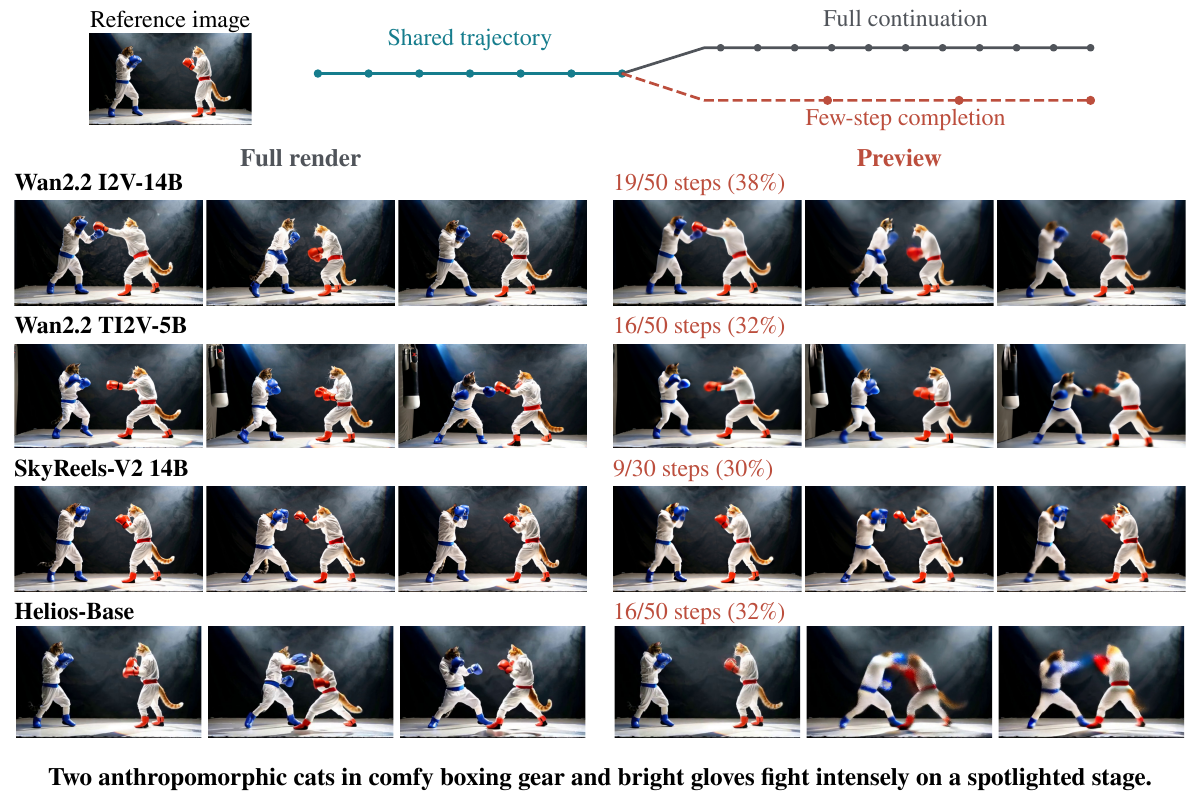}
    \caption{\textbf{Image-conditioned previews and full renders.}
    Paired outputs share the reference image, prompt, and seed.
    Percentages denote sampling-step ratios, not runtime ratios.}
    \label{fig:motivation_observation_i2v}
\end{figure}
\endgroup

\clearpage
\section{Pipeline Algorithm}
\label{app:pipeline_algorithm}

Algorithm~\ref{alg:precogqa} details the preview-guided correction loop,
including condition validation, fallback selection, and full rendering.

\input{PreCoGQA}
\FloatBarrier

\input{PreCoGQA_Details}

\FloatBarrier
\input{Experiments_Human}
\FloatBarrier

\input{Experimental_Details}

\FloatBarrier
\input{Metric_Breakdown}

\FloatBarrier
\input{Additional_Ablations}
\FloatBarrier

\input{Experiments_Generalization}
\FloatBarrier



%% file: PreCoGQA.tex
\begin{algorithm}[H]
\caption{\precog{}: Preview-Guided Corrective Generation}
\label{alg:precogqa}
\begin{algorithmic}[1]
\Require Original prompt $P_0$, preview budget $N\geq1$, editing configuration
         $\Theta=(W_{\mathrm{edit}},W_{\mathrm{cf}},\beta,\gamma)$
\Ensure Full-render video $X^*$
\State $(P_{\mathrm{src}},\mathcal{L}_0)\gets\Call{PlanningAgent.Design}{P_0}$
\Comment{Adapting GENMAC}
\State $\mathcal{Q}\gets\Call{PlanningAgent.Requirements}{P_0}$
\State $(V_0,\mathcal{C}_0,\mathcal{B}_0)
       \gets\Call{GenerateGuidedPreview}{P_{\mathrm{src}},\mathcal{L}_0}$
\State $P_0^{\mathrm{gen}}\gets P_{\mathrm{src}}$, $\mathcal{H}_0\gets\emptyset$,
       $i^*\gets0$
\For{$i=0,\ldots,N-1$}
    \State $\mathcal{R}_i\gets\Call{EvaluationAgent}{V_i,\mathcal{Q}}$
    \State $(S_i,\mathcal{F}_i)\gets\Call{AggregateVerdicts}{\mathcal{R}_i}$
    \Comment{$S_i$: number of satisfied requirements}
    \State $\mathcal{H}_{i+1}\gets\mathcal{H}_i\cup
    \{(i,V_i,P_i^{\mathrm{gen}},\mathcal{R}_i,S_i,\mathcal{C}_i,\mathcal{B}_i)\}$
    \If{$S_i>S_{i^*}$}
        \State $i^*\gets i$ \Comment{Keep the earlier branch on ties}
    \EndIf
    \If{$\mathcal{F}_i=\emptyset$ \textbf{or} $i=N-1$}
        \State \textbf{break}
    \EndIf
    \State $\mathcal{E}_i\gets
    \Call{PlanAllFailedGroups}{\mathcal{Q},\mathcal{R}_{i^*},\mathcal{H}_{i+1}}$
    \If{all repair strategies for the current failure state are exhausted}
        \State \textbf{break} \Comment{Select the branch with highest requirement coverage}
    \EndIf
    \State $L_i\gets\Call{RefinementAgent.Refine}{V_{i^*},\mathcal{E}_i,\mathcal{H}_{i+1}}$
    \State $P_i^{\mathrm{tar}}\gets\Call{RefinementAgent.Integrate}{P_0,\mathcal{Q},L_i,\mathcal{E}_i}$
    \State $P_i^{\mathrm{cf}}\gets\Call{RefinementAgent.Counterfactual}{P_i^{\mathrm{tar}},\mathcal{R}_{i^*},\mathcal{E}_i}$
    \Comment{Validated target--counterfactual contrast}
    \If{no valid condition pair is obtained}
        \State \textbf{break} \Comment{Select the branch with highest requirement coverage}
    \EndIf
    \State $P_{i+1}^{\mathrm{gen}}\gets P_i^{\mathrm{tar}}$
    \State $(V_{i+1},\mathcal{C}_{i+1},\mathcal{B}_{i+1})\gets
    \Call{\preedit{}Preview}{P_i^{\mathrm{tar}},P_i^{\mathrm{cf}},\mathcal{C}_{i^*};\Theta}$
\EndFor
\State $X^*\gets\Call{RenderFullVideo}{P_{i^*}^{\mathrm{gen}},\mathcal{B}_{i^*}}$
\Comment{Complete the selected trajectory with the native continuation}
\State \Return $X^*$
\end{algorithmic}
\end{algorithm}

%% file: PreCoGQA_Details.tex
\section{\preqa{}: Planning, Evaluation, and Refinement Agents}
\label{app:precogqa_details}

\paragraph{Planning Agent.}
Requirements receive stable identifiers and are grouped by canonical subject.
Each event requirement retains its participants, direction, manner, and temporal structure in one assertion, preventing partial satisfaction from counting as success.
Source initialization settings are given in Appendix~\ref{app:experimental_details}.

\paragraph{Evaluation Agent.}
The agent assesses all 81 preview frames.
Failed physical or temporal requirements receive a supplementary check using three frames near the beginning and three near the end.
This check can overturn a failure only when it establishes the complete event; endpoint changes alone cannot establish actions requiring intermediate motion.

\paragraph{Refinement Agent.}
The agent uses previous assessments and executed conditions to rephrase failed predicates, explicitly describe required events, or bundle related requirements.
Bundles may include earlier failures from the selected groups to address alternating failures.
Repeated conditions and unchanged outputs inform subsequent choices.

\paragraph{Condition construction and validation.}
The counterfactual condition is distinct from the generator's default CFG negative prompt.
Validation checks target fidelity, evidence for the replacements, and a contrast restricted to the selected attributes under a shared context.
Local correction descriptions allow up to two rewrites, followed by at most three validation rounds for scene and condition-pair construction.
Revising failure values alone leaves the validated target condition fixed.

%% file: Experiments_Human.tex
\section{Human Preference Study}
\label{sec:exp_human}

We report a human preference evaluation comparing \precog{} against VQQA, VSR, and GENMAC in blinded pairs.
The evaluation includes 96 distinct videos, each assessed by all four raters.
Each pair uses the same prompt and matched output specifications. Left--right placement is randomized
and balanced for each baseline. Visual quality, prompt alignment, and physical naturalness are judged
separately, with a tie option for each criterion. Both initial and edited selections are included.

\begin{figure}[htbp]
    \centering
    \includegraphics[width=\linewidth]{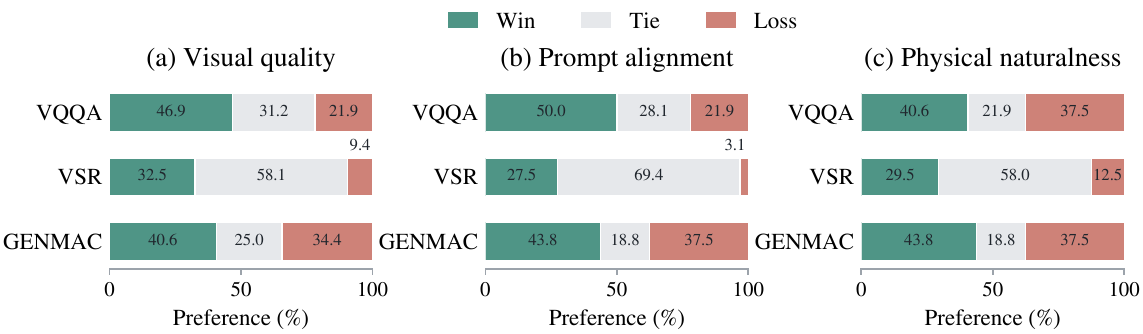}
    \caption{\textbf{Human preference.} \precog{} win/tie/loss percentages for
    visual quality, prompt alignment, and physical naturalness.}
    \label{fig:human_preference}
\end{figure}

Figure~\ref{fig:human_preference} shows more wins than losses for \precog{}
in visual quality and prompt alignment against all three baselines.
The largest margins occur against VQQA, with win/loss shares of
46.9\%/21.9\% for visual quality and 50.0\%/21.9\% for prompt alignment.
\precog{} also outperforms GENMAC in prompt alignment, with 43.8\% wins
versus 37.5\% losses. Against VSR, ties account for 58.0--69.4\% of judgments
across the three criteria. The paired outputs were therefore often judged comparable,
although decisions expressing a preference favored
\precog{} in every criterion.

Physical naturalness likewise favors \precog{}, although the margins against
VQQA and GENMAC are modest. Win/loss shares are 40.6\%/37.5\% against VQQA
and 43.8\%/37.5\% against GENMAC. The preference for \precog{} thus extends
across all three criteria, with stronger gains in visual quality and prompt
alignment against VQQA than in physical naturalness.

%% file: Experimental_Details.tex
\section{Detailed Experimental Settings}
\label{app:experimental_details}

\subsection{Baseline Configurations}
We compare \precog{} with direct generation, inference-time search
(full best-of-$N$ and CachedSearch), and agentic generation (VQQA, VSR, and GENMAC).
All methods start from the original prompt and use Wan2.2-T2V-A14B with
matched resolution and frame count. Methods requiring online scoring or
feedback use Qwen3.8-27B. The configurations below describe our Wan2.2
implementations, including their generation budgets and output selection rules.

\paragraph{Direct generation.}
The native backbone generates one full video from the original prompt using
the $40$-step schedule, without feedback or candidate selection.

\paragraph{Full best-of-$N$.}
We generate four full videos from the original prompt using a fixed list of
distinct seeds. We use the same candidate scoring and selection rule as our
CachedSearch implementation: Qwen rates each video against the original prompt
on a 0--100 scale, and we deliver the highest-scoring candidate, breaking ties
by candidate order.

\paragraph{CachedSearch.}
Our Wan2.2 adaptation of CachedSearch~\citep{cachedsearch} explores eight
seeds using cached denoiser computations and the Qwen scoring and selection
rule described above.
The two search methods share the first four candidate seeds.
We use a cache threshold of $0.10$, with five warmup and five cooldown steps.
After selection, the winning seed is regenerated with caching disabled to
produce the delivered full video. Its inference cost includes all eight
exploration runs, online scoring, and this final regeneration.

\paragraph{VQQA.}
Our VQQA~\citep{vqqa} implementation evaluates each full video with a
global score, generates questions about alignment and quality, and uses the
answers to refine the prompt before full regeneration.
We allow an initial generation and at most four refinements.
The loop stops when the global score reaches $100$ or the best score fails
to improve over three refinements. The highest-scoring completed video is
delivered, with earlier candidates preferred in ties.

\paragraph{VSR.}
We adapt the Visual Self-Refine baseline described in VISTA~\citep{vista}
to Wan2.2, following the paradigm of iterative feedback and refinement of
Self-Refine~\citep{selfrefine}. Starting from the original prompt, Qwen inspects each video and
rewrites the prompt for the next full generation.
We use four generations in total, retain seed $17$ across iterations, and
deliver the final video without early stopping or selection of the best candidate.

\paragraph{GENMAC.}
Our Wan2.2 implementation of GENMAC~\citep{genmac} retains multi-agent
layout planning, verification, and correction, together with object-weighted
layout guidance during generation. Guidance uses layers $4$, $8$, and $12$
over the first $10$ denoising steps, with scale $10$ and one guidance update
per step. We allow an initial generation and at most two corrections, using
seed $17$ throughout. The loop stops when verification requests no further
revision or the budget is exhausted, and delivers the latest video.
The controlled \preqa{} plus regeneration comparison is specified separately
in Section~\ref{sec:exp_ablation}.

\subsection{\precog{} Configuration}

\paragraph{Source initialization.}
Qwen3.8-27B produces the initial layout and source description, including the scene, background, and object-token alignment, following GENMAC's design procedure.
Layout guidance uses layers 4, 8, and 12 during the first ten native denoising steps, with guidance scale 10.
These settings are fixed during correction, and source generation is not retried with another seed.
Initialization time includes planning and layout-guidance computation.

\paragraph{Generation and editing.}
The main configuration uses Wan2.2-T2V-A14B with its native $K=40$-step
schedule. Source generation and continuation after editing use UniPC, while
the source-relative displacement is integrated with Euler as specified
in Eq.~\ref{eq:edited_solver_step}. We set $g_{\mathrm{src}}=4$,
$g_{\mathrm{tar}}=10$, and $\gamma=0.65$. The edit interval is
$k_a=1$ to $k_b=14$ in our zero-based notation, corresponding to
the one-based inclusive interval $[2,14]$ (steps 2--14).
Solver history is reset at the start of editing, on returning to
ordinary denoising, and at the boundaries of the sparse schedule.
We use blockwise soft projection with
$\alpha_{\max}=1.5$ and $\beta=0.75$. Each block spans $3\times30\times52$
positions in the temporal and spatial latent dimensions, including all
channels. The projection window is $W_{\mathrm{cf}}=[1,5)$ in zero-based
notation, corresponding to denoising steps 2--5. The remaining steps in
$W_{\mathrm{edit}}=[1,14)$ use regular source-relative updates.
No additional contrastive gain or coefficient smoothing is used.
Attention scaling, KV preservation, and spatial masks are disabled.

\paragraph{Cached trajectory details.}
For the initial candidate, cached states at guided steps include initialization
guidance. We retain the pre- and post-update states and guidance displacements
to replay the source without recomputing layout gradients. Initial velocities
are the generator's CFG predictions under the source condition and
$g_{\mathrm{src}}$. For corrected candidates, we cache the actual states and
applied editing velocities, together with the recoverable solver state.

\paragraph{Geometry of soft projection.}
\label{app:projection_geometry}
For one block, omit the iteration and step indices and ignore $\epsilon$.
The projected update $\mathbf p$ is the Euclidean projection of $\mathbf d$
onto the convex segment $\mathcal C=\{\lambda\mathbf a:
0\leq\lambda\leq\alpha_{\max}\}$. For any $\mathbf d^*\in\mathcal C$,
\begin{equation}
\|\mathbf u-\mathbf d^*\|_2^2
\leq\|\mathbf d-\mathbf d^*\|_2^2
-(1-\beta^2)\|\mathbf d-\mathbf p\|_2^2.
\label{eq:projection_contraction}
\end{equation}
This follows from the projection inequality
$\langle\mathbf d-\mathbf p,\mathbf d^*-\mathbf p\rangle\leq0$
and $\mathbf u=\mathbf p+\beta(\mathbf d-\mathbf p)$.
The bound concerns corrections within $\mathcal C$ and does not guarantee
semantic alignment when the counterfactual direction is inaccurate.

\paragraph{Preview schedule.}
We retain the first $K_p=14$ steps of the native trajectory and use a $14+2+2$ preview
schedule, giving $K_{\mathrm{pre}}=18$. After this initial segment, two uniformly
spaced sigma transitions bridge to the high-/low-noise expert switch,
followed by two transitions to zero noise. Full rendering restores the
selected branch at the end of the edit interval and completes the native
continuation, as described in Section~\ref{sec:preview_render}.

\paragraph{Feedback configuration.}
We allow at most four editing rounds after the initial preview,
corresponding to $N=5$ evaluated previews. Only previews receive online semantic assessment.
The selected result is subsequently rendered as a full video.
The assessment, group selection, and condition construction protocols are
detailed in Appendix~\ref{app:precogqa_details}.

\subsection{Evaluation and Timing}

\paragraph{Metric computation.}
T2V-Comp. denotes the macro average across seven categories over available outputs in
each category. VBench is the equal mean of six quality dimensions on a
common 0--100 scale, rather than the official VBench total.
VS2 denotes the mean VideoScore2 soft text-to-video alignment score,
computed using the upstream rule that weights rating tokens by confidence.
Further metric definitions and hard/soft score breakdowns are given in
Appendix~\ref{app:metric_breakdown}. Offline outcome scores are never
used for online candidate selection.

%% file: Metric_Breakdown.tex
\section{Metric Definitions and Baseline Quality}
\label{app:metric_breakdown}

Table~\ref{tab:main_results} and the
tables below for individual dimensions compare methods on matched prompts with completed
outputs. Prompts with missing outputs are excluded consistently across
methods. Missing outputs are not assigned zero.

\paragraph{VBench.}
The six dimensions are subject consistency (SC), background consistency (BC),
motion smoothness (MS), aesthetic quality (AQ), imaging quality (IQ), and
dynamic degree (DD). We average each dimension over available
final videos. For display, all dimensions except IQ are multiplied by
100. IQ already uses a 0--100 scale. The main table reports VBench: the unweighted arithmetic mean of
SC, BC, MS, AQ, IQ, and DD on this common 0--100 scale. This is the
requested average across six dimensions on the CompBench subset, without the
normalization or weighting of the official total score. Temporal flickering
is retained only as an ancillary measurement and is excluded from this
average. No semantic dimensions enter VBench.

\input{vbench_real}

\paragraph{T2V-CompBench.}
The seven categories are consistent attribute binding (CAB), dynamic attribute
binding (DAB), spatial relationships (SR), motion binding (MB), action binding
(AB), object interactions (OI), and generative numeracy (GN). Each video is evaluated
in its assigned category. T2V-Comp. averages the seven category means with equal weight
and scales the result by 100.

\input{compbench_real}

\paragraph{VideoScore2.}
The main table reports the mean soft text-to-video alignment score. The local
upstream implementation defines this score as the most probable rating token's
value multiplied by its probability normalized over tokens 1--5, rounded
to four decimals per video. It is not an expected rating over all five
tokens. Table~\ref{tab:main_results} uses matched prompts across methods.
Results for each video, coverage, sample variances, and source hashes are
retained in the accompanying data export.

\paragraph{Inference time.}
We report end-to-end inference time, including generation, caching,
online scoring or feedback, and final rendering.
Baseline timing results retain their cohorts with available outputs rather than
the matched quality subset used for quality and alignment metrics.

\input{quality_efficiency_tradeoff}

%% file: vbench_real.tex
\begin{table}[htbp]
\centering
\caption{\textbf{VBench scores for each dimension.} All methods use the same prompts as Table~\ref{tab:main_results}. Scores are on a 0--100 scale. Avg. is the mean across six dimensions. Bold and underlined values indicate the best and second-best results, respectively.}
\begin{tabular*}{\linewidth}{@{\extracolsep{\fill}}lrrrrrrr@{}}
\toprule
Method & SC & BC & MS & AQ & IQ & DD & Avg. \\
\midrule
Direct generation & 92.18 & 94.02 & 98.21 & 57.99 & 68.72 & \textbf{81.25} & 82.06 \\
Full best-of-N (N=4) & 92.32 & 94.60 & \underline{98.47} & 58.44 & 68.92 & 75.00 & 81.29 \\
CachedSearch (N=8) & 92.09 & 94.44 & 98.27 & 58.22 & 68.38 & \textbf{81.25} & 82.11 \\
VQQA & 93.08 & \underline{94.61} & 98.45 & 58.34 & 67.96 & 68.75 & 80.20 \\
VSR & 92.69 & 94.26 & 98.20 & 57.27 & 65.13 & 75.00 & 80.43 \\
GENMAC & \textbf{93.92} & 94.59 & \textbf{98.51} & \underline{60.25} & \textbf{71.15} & \underline{78.12} & \underline{82.76} \\
\textbf{\precog{}} & \underline{93.35} & \textbf{95.80} & 98.38 & \textbf{64.32} & \underline{70.95} & \underline{78.12} & \textbf{83.49} \\
\bottomrule
\end{tabular*}
\end{table}

%% file: compbench_real.tex
\begin{table}[htbp]
\centering
\caption{\textbf{T2V-CompBench category scores.} Prompts match Table~\ref{tab:main_results}. Category means use a 0--100 scale. Bold and underlined values mark the best and second-best results.}
\begin{tabular*}{\linewidth}{@{\extracolsep{\fill}}lrrrrrrr@{}}
\toprule
Method & CAB & DAB & SR & MB & AB & OI & GN \\
\midrule
Direct generation & \underline{97.14} & 3.20 & 54.23 & \underline{37.95} & 69.63 & 83.33 & \underline{41.41} \\
Full best-of-N (N=4) & \textbf{97.62} & 1.28 & 56.46 & 28.48 & 71.11 & 90.74 & 22.66 \\
CachedSearch (N=8) & \underline{97.14} & \underline{16.00} & \textbf{68.82} & 21.73 & 69.63 & 90.74 & 21.09 \\
VQQA & 86.19 & 1.28 & \underline{68.17} & 22.71 & 65.93 & 83.33 & 35.94 \\
VSR & 95.24 & \textbf{25.60} & 61.35 & 35.85 & \underline{85.19} & \underline{99.07} & 30.47 \\
GENMAC & 85.24 & 1.15 & 67.29 & 28.82 & 63.70 & \textbf{100.00} & 24.22 \\
\textbf{\precog{}} & 96.67 & 13.44 & 54.32 & \textbf{42.00} & \textbf{90.04} & 90.74 & \textbf{48.75} \\
\bottomrule
\end{tabular*}
\end{table}

%% file: quality_efficiency_tradeoff.tex
\begin{figure}[!htbp]
    \centering
    \includegraphics[width=\linewidth]{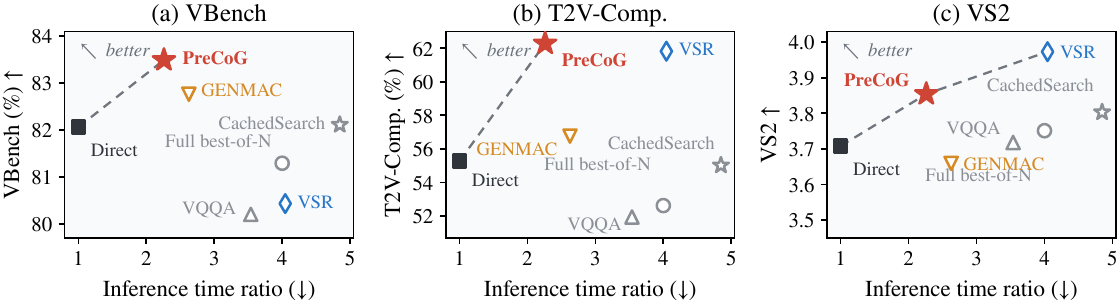}
    \caption{\textbf{Quality and alignment versus inference cost.}
    Points show the results in Table~\ref{tab:main_results}, with inference time
    normalized to direct generation. Dashed lines connect Pareto-optimal methods
    among the compared results; higher scores and lower costs are preferred.}
    \label{fig:quality_efficiency_tradeoff}
\end{figure}

%% file: Additional_Ablations.tex
\section{Additional Ablation Analyses}
\label{app:additional_ablation}

%
%
\subsection{VLM Ablation}
\label{app:vlm_ablation}

Table~\ref{tab:vlm_ablation} compares Qwen3-VL (4B and 30B)~\citep{qwen3vl},
Qwen3.8-27B~\citep{qwen38}, Gemini 3.1 Pro~\citep{gemini31pro}, and GPT-4.1~\citep{gpt41} as the \preqa{} model.
We fix Wan2.2-T2V-A14B, the preview schedule, four correction rounds, and default soft projection.
Prompts, source seeds, frame inputs, and QA instructions are matched; all model-dependent QA calls use the selected model.
The original requirement set and independent final-video evaluators remain fixed across variants.

\input{vlm_ablation}

All tested VLMs except Qwen VL 3 4B improve all three metrics over direct generation; the exception may reflect the smaller model's limited feedback capability.
Across these models, T2V-Comp. gains range from 4.41 to 7.01 percentage points, accompanied by improvements in VBench and VS2.
These results support the use of \precog{} with multiple sufficiently capable feedback models.

\FloatBarrier

%% file: vlm_ablation.tex
\begin{table}[H]
\centering

\setlength{\tabcolsep}{2pt}
\caption{\textbf{\preqa{} model ablation.} $\Delta$ denotes change from direct generation (Table~\ref{tab:main_results}). Bold values indicate the best results.}
\label{tab:vlm_ablation}
\begin{tabular*}{\linewidth}{@{\extracolsep{\fill}}lr@{\extracolsep{0pt}\hspace{4pt}}>{\color{comparisontext}}r@{\extracolsep{\fill}\hspace{10pt}}r@{\extracolsep{0pt}\hspace{4pt}}>{\color{comparisontext}}r@{\extracolsep{\fill}\hspace{10pt}}r@{\extracolsep{0pt}\hspace{4pt}}>{\color{comparisontext}}r@{}}
\toprule
\preqa{} model & \multicolumn{2}{c}{VBench (\% $\uparrow$)} & \multicolumn{2}{c}{T2V-Comp. (\% $\uparrow$)} & \multicolumn{2}{c}{VS2 ($\uparrow$)} \\
\cmidrule(lr){2-3}\cmidrule(lr){4-5}\cmidrule(lr){6-7}
 & Value & $\Delta$ & Value & $\Delta$ & Value & $\Delta$ \\
\midrule
Direct generation & 82.06 & 0.00 & 55.27 & 0.00 & 3.708 & 0.000 \\
\midrule
Qwen VL 3 4B & 81.72 & \deltaloss{-0.34} & 51.86 & \deltaloss{-3.41} & 3.584 & \deltaloss{-0.124} \\
Qwen VL 3 30B & 83.31 & \deltagain{+1.25} & 59.68 & \deltagain{+4.41} & 3.827 & \deltagain{+0.119} \\
Qwen3.8-27B & 83.49 & \deltagain{+1.43} & \textbf{62.28} & \deltagain{+7.01} & 3.854 & \deltagain{+0.146} \\
Gemini 3.1 Pro & 83.46 & \deltagain{+1.40} & 62.08 & \deltagain{+6.81} & 3.901 & \deltagain{+0.193} \\
GPT-4.1 & \textbf{83.62} & \deltagain{+1.56} & 61.73 & \deltagain{+6.46} & \textbf{3.918} & \deltagain{+0.210} \\
\bottomrule
\end{tabular*}
\end{table}

%% file: Experiments_Generalization.tex
\subsection{Backbone and Task Ablations}
\label{sec:exp_generalization}

Table~\ref{tab:generalization} compares \precog{} with direct native generation across four T2V and two I2V configurations.
All configurations use Qwen3.8-27B as the \preqa{} model.
Each pair uses matched prompts, initial noise seeds, output specifications, and input images for I2V.
Metrics follow Table~\ref{tab:main_results}.

\input{generalization}

\precog{} improves all three metrics across the six backbone--task configurations in Table~\ref{tab:generalization}.
T2V-Comp. increases by 4.80--7.01 percentage points, while VBench and VS2 increase by 0.65--1.43 percentage points and 0.130--0.190, respectively.
For I2V, T2V-Comp. gains of 4.80 and 5.20 percentage points show that correction remains beneficial when generation is conditioned on an input image.
These comparisons support the applicability of preview-guided correction across the tested backbones and both conditioning tasks.

%% file: generalization.tex
\begin{table}[htbp]
\centering

\setlength{\tabcolsep}{2pt}
\caption{\textbf{\precog{} across backbones and tasks.} $\Delta$ denotes change from direct generation.}
\label{tab:generalization}
\begin{tabular*}{\linewidth}{@{\extracolsep{\fill}}lccccccccc@{}}
\toprule
Backbone & \multicolumn{3}{c}{VBench (\% $\uparrow$)} & \multicolumn{3}{c}{T2V-Comp. (\% $\uparrow$)} & \multicolumn{3}{c}{VS2 ($\uparrow$)} \\
\cmidrule(lr){2-4}\cmidrule(lr){5-7}\cmidrule(lr){8-10}
 & Direct & \precog{} & $\Delta$ & Direct & \precog{} & $\Delta$ & Direct & \precog{} & $\Delta$ \\
\midrule
\multicolumn{10}{l}{\textit{T2V}} \\
Wan2.2-T2V-A14B & 82.06 & 83.49 & \deltagain{+1.43} & 55.27 & 62.28 & \deltagain{+7.01} & 3.708 & 3.854 & \deltagain{+0.146} \\
Wan2.2-TI2V-5B & 79.80 & 81.05 & \deltagain{+1.25} & 48.60 & 54.20 & \deltagain{+5.60} & 3.420 & 3.610 & \deltagain{+0.190} \\
SkyReels-V2-T2V-14B & 81.20 & 82.35 & \deltagain{+1.15} & 52.80 & 58.90 & \deltagain{+6.10} & 3.580 & 3.750 & \deltagain{+0.170} \\
Helios-Base & 80.70 & 81.65 & \deltagain{+0.95} & 50.40 & 55.70 & \deltagain{+5.30} & 3.490 & 3.650 & \deltagain{+0.160} \\
\midrule
\multicolumn{10}{l}{\textit{I2V}} \\
Wan2.2-TI2V-5B & 82.10 & 82.85 & \deltagain{+0.75} & 54.30 & 59.10 & \deltagain{+4.80} & 3.600 & 3.740 & \deltagain{+0.140} \\
Wan2.2-I2V-A14B & 83.40 & 84.05 & \deltagain{+0.65} & 58.20 & 63.40 & \deltagain{+5.20} & 3.760 & 3.890 & \deltagain{+0.130} \\
\bottomrule
\end{tabular*}
\end{table}

%% file: arxiv.bbl
\begin{thebibliography}{27}
\providecommand{\natexlab}[1]{#1}
\providecommand{\url}[1]{\texttt{#1}}
\expandafter\ifx\csname urlstyle\endcsname\relax
  \providecommand{\doi}[1]{doi: #1}\else
  \providecommand{\doi}{doi: \begingroup \urlstyle{rm}\Url}\fi

\bibitem[Bai et~al.(2025)Bai, Cai, Chen, Chen, Chen, Cheng, Deng, Ding, Gao,
  Ge, Ge, Guo, Huang, Huang, Huang, Hui, Jiang, Li, Li, Li, Li, Lin, Lin, Liu,
  Liu, Liu, Liu, Liu, Liu, Lu, Luo, Lv, Men, Meng, Ren, Ren, Song, Sun, Tang,
  Tu, Wan, Wang, Wang, Wang, Wang, Xie, Xu, Xu, Xu, Yang, Yang, Yang, Yang, Yu,
  Zhang, Zhang, Zhang, Zheng, Zhong, Zhou, Zhou, Zhou, Zhu, and Zhu]{qwen3vl}
Shuai Bai, Yuxuan Cai, Ruizhe Chen, Keqin Chen, Xionghui Chen, Zesen Cheng,
  Lianghao Deng, Wei Ding, Chang Gao, Chunjiang Ge, Wenbin Ge, Zhifang Guo,
  Qidong Huang, Jie Huang, Fei Huang, Binyuan Hui, Shutong Jiang, Zhaohai Li,
  Mingsheng Li, Mei Li, Kaixin Li, Zicheng Lin, Junyang Lin, Xuejing Liu,
  Jiawei Liu, Chenglong Liu, Yang Liu, Dayiheng Liu, Shixuan Liu, Dunjie Lu,
  Ruilin Luo, Chenxu Lv, Rui Men, Lingchen Meng, Xuancheng Ren, Xingzhang Ren,
  Sibo Song, Yuchong Sun, Jun Tang, Jianhong Tu, Jianqiang Wan, Peng Wang,
  Pengfei Wang, Qiuyue Wang, Yuxuan Wang, Tianbao Xie, Yiheng Xu, Haiyang Xu,
  Jin Xu, Zhibo Yang, Mingkun Yang, Jianxin Yang, An~Yang, Bowen Yu, Fei Zhang,
  Hang Zhang, Xi~Zhang, Bo~Zheng, Humen Zhong, Jingren Zhou, Fan Zhou, Jing
  Zhou, Yuanzhi Zhu, and Ke~Zhu.
\newblock {Qwen3-VL Technical Report}, 2025.
\newblock URL \url{https://arxiv.org/abs/2511.21631}.

\bibitem[Chen et~al.(2025)Chen, Lin, Yang, Lin, Zhu, Fan, Zhang, Chen, Chen,
  Ma, Xiong, Wang, Pang, Kang, Xu, Jin, Liang, Song, Zhao, Xu, Qiu, Li, Fei,
  Li, and Zhou]{skyreelsv2}
Guibin Chen, Dixuan Lin, Jiangping Yang, Chunze Lin, Junchen Zhu, Mingyuan Fan,
  Hao Zhang, Sheng Chen, Zheng Chen, Chengcheng Ma, Weiming Xiong, Wei Wang,
  Nuo Pang, Kang Kang, Zhiheng Xu, Yuzhe Jin, Yupeng Liang, Yubing Song, Peng
  Zhao, Boyuan Xu, Di~Qiu, Debang Li, Zhengcong Fei, Yang Li, and Yahui Zhou.
\newblock {SkyReels-V2: Infinite-length Film Generative Model}, 2025.
\newblock URL \url{https://arxiv.org/abs/2504.13074}.

\bibitem[Chen et~al.(2026)Chen, Chen, Li, and Mao]{flowanchor}
Ze~Chen, Lan Chen, Yuanhang Li, and Qi~Mao.
\newblock Flowanchor: Stabilizing the editing signal for inversion-free video
  editing, 2026.
\newblock URL \url{https://arxiv.org/abs/2604.22586}.

\bibitem[Choi et~al.(2025)Choi, Sharan, Goel, Shah, and Chinchali]{neuse}
Minkyu Choi, S~P Sharan, Harsh Goel, Sahil Shah, and Sandeep Chinchali.
\newblock We'll fix it in post: Improving text-to-video generation with
  neuro-symbolic feedback, 2025.
\newblock URL \url{https://arxiv.org/abs/2504.17180}.

\bibitem[{Google DeepMind}(2026)]{gemini31pro}
{Google DeepMind}.
\newblock {Gemini 3.1 Pro Model Card}.
\newblock \url{https://deepmind.google/models/model-cards/gemini-3-1-pro/},
  February 2026.

\bibitem[He et~al.(2025)He, Jiang, Nie, Liu, Jiang, Su, Ma, Lin, Ye, Lu, Wu,
  Schneider, Do, Li, Jia, Zhang, Cheng, Wang, Zhou, Lin, Zhang, Zhang, Huang,
  and Chen]{videoscore2}
Xuan He, Dongfu Jiang, Ping Nie, Minghao Liu, Zhengxuan Jiang, Mingyi Su,
  Wentao Ma, Junru Lin, Chun Ye, Yi~Lu, Keming Wu, Benjamin Schneider, Quy~Duc
  Do, Zhuofeng Li, Yiming Jia, Yuxuan Zhang, Guo Cheng, Haozhe Wang,
  Wangchunshu Zhou, Qunshu Lin, Yuanxing Zhang, Ge~Zhang, Wenhao Huang, and
  Wenhu Chen.
\newblock {VideoScore2}: Think before you score in generative video evaluation,
  2025.
\newblock URL \url{https://arxiv.org/abs/2509.22799}.

\bibitem[Huang et~al.(2026)Huang, Huang, Ning, Lin, Wang, and Liu]{genmac}
Kaiyi Huang, Yukun Huang, Xuefei Ning, Zinan Lin, Yu~Wang, and Xihui Liu.
\newblock Genmac: compositional text-to-video generation with multi-agent
  collaboration.
\newblock In \emph{Proceedings of the Fortieth AAAI Conference on Artificial
  Intelligence and Thirty-Eighth Conference on Innovative Applications of
  Artificial Intelligence and Sixteenth Symposium on Educational Advances in
  Artificial Intelligence}, AAAI'26/IAAI'26/EAAI'26. AAAI Press, 2026.
\newblock ISBN 978-1-57735-906-7.
\newblock \doi{10.1609/aaai.v40i7.37418}.
\newblock URL \url{https://doi.org/10.1609/aaai.v40i7.37418}.

\bibitem[Huang et~al.(2024)Huang, He, Yu, Zhang, Si, Jiang, Zhang, Wu, Jin,
  Chanpaisit, Wang, Chen, Wang, Lin, Qiao, and Liu]{vbench}
Ziqi Huang, Yinan He, Jiashuo Yu, Fan Zhang, Chenyang Si, Yuming Jiang, Yuanhan
  Zhang, Tianxing Wu, Qingyang Jin, Nattapol Chanpaisit, Yaohui Wang, Xinyuan
  Chen, Limin Wang, Dahua Lin, Yu~Qiao, and Ziwei Liu.
\newblock {VBench}: Comprehensive benchmark suite for video generative models.
\newblock In \emph{Proceedings of the IEEE/CVF Conference on Computer Vision
  and Pattern Recognition}, pp.\  21807--21818, 2024.
\newblock URL \url{https://arxiv.org/abs/2311.17982}.

\bibitem[Kulikov et~al.(2024)Kulikov, Kleiner, Huberman-Spiegelglas, and
  Michaeli]{flowedit}
Vladimir Kulikov, Matan Kleiner, Inbar Huberman-Spiegelglas, and Tomer
  Michaeli.
\newblock Flowedit: Inversion-free text-based editing using pre-trained flow
  models, 2024.
\newblock URL \url{https://arxiv.org/abs/2412.08629}.

\bibitem[Lee et~al.(2024)Lee, Yoon, Cho, and Bansal]{videorepair}
Daeun Lee, Jaehong Yoon, Jaemin Cho, and Mohit Bansal.
\newblock Self-correcting text-to-video generation with misalignment detection
  and localized refinement, 2024.
\newblock URL \url{https://arxiv.org/abs/2411.15115}.

\bibitem[Li et~al.(2025{\natexlab{a}})Li, Yang, Song, and Zhang]{flowdirector}
Guangzhao Li, Yanming Yang, Chenxi Song, and Chi Zhang.
\newblock Flowdirector: Training-free flow steering for precise text-to-video
  editing, 2025{\natexlab{a}}.
\newblock URL \url{https://arxiv.org/abs/2506.05046}.

\bibitem[Li et~al.(2025{\natexlab{b}})Li, Xie, Wu, Zhang, and Wang]{five}
Minghan Li, Chenxi Xie, Yichen Wu, Lei Zhang, and Mengyu Wang.
\newblock Five: A fine-grained video editing benchmark for evaluating emerging
  diffusion and rectified flow models, 2025{\natexlab{b}}.
\newblock URL \url{https://arxiv.org/abs/2503.13684}.

\bibitem[Liu et~al.(2025)Liu, Wang, Cai, Zhang, Zhan, and Duan]{videot1}
Fangfu Liu, Hanyang Wang, Yimo Cai, Kaiyan Zhang, Xiaohang Zhan, and Yueqi
  Duan.
\newblock Video-t1: Test-time scaling for video generation, 2025.
\newblock URL \url{https://arxiv.org/abs/2503.18942}.

\bibitem[Long et~al.(2025)Long, Wan, Nakhost, Lee, Pfister, and Arik]{vista}
Do~Xuan Long, Xingchen Wan, Hootan Nakhost, Chen-Yu Lee, Tomas Pfister, and
  Sercan~O. Arik.
\newblock Vista: A test-time self-improving video generation agent, 2025.
\newblock URL \url{https://arxiv.org/abs/2510.15831}.

\bibitem[Madaan et~al.(2023)Madaan, Tandon, Gupta, Hallinan, Gao, Wiegreffe,
  Alon, Dziri, Prabhumoye, Yang, Gupta, Majumder, Hermann, Welleck,
  Yazdanbakhsh, and Clark]{selfrefine}
Aman Madaan, Niket Tandon, Prakhar Gupta, Skyler Hallinan, Luyu Gao, Sarah
  Wiegreffe, Uri Alon, Nouha Dziri, Shrimai Prabhumoye, Yiming Yang, Shashank
  Gupta, Bodhisattwa~Prasad Majumder, Katherine Hermann, Sean Welleck, Amir
  Yazdanbakhsh, and Peter Clark.
\newblock {Self-Refine}: Iterative refinement with self-feedback, 2023.
\newblock URL \url{https://arxiv.org/abs/2303.17651}.

\bibitem[{OpenAI}(2025)]{gpt41}
{OpenAI}.
\newblock Introducing {GPT-4.1} in the {API}.
\newblock \url{https://openai.com/index/gpt-4-1/}, April 2025.

\bibitem[Oshima et~al.(2025)Oshima, Suzuki, Matsuo, and Furuta]{latentbeam}
Yuta Oshima, Masahiro Suzuki, Yutaka Matsuo, and Hiroki Furuta.
\newblock Inference-time text-to-video alignment with diffusion latent beam
  search, 2025.
\newblock URL \url{https://arxiv.org/abs/2501.19252}.

\bibitem[Qian et~al.(2024)Qian, Cai, Pan, Li, Yao, Sun, and Mei]{masf}
Yurui Qian, Qi~Cai, Yingwei Pan, Yehao Li, Ting Yao, Qibin Sun, and Tao Mei.
\newblock Boosting diffusion models with moving average sampling in frequency
  domain.
\newblock In \emph{Proceedings of the IEEE/CVF Conference on Computer Vision
  and Pattern Recognition}, pp.\  8911--8920, 2024.
\newblock URL \url{https://arxiv.org/abs/2403.17870}.

\bibitem[{Qwen Team}(2026)]{qwen38}
{Qwen Team}.
\newblock {Qwen3.8}: Official model repository.
\newblock \url{https://github.com/QwenLM/Qwen3.8}, 2026.
\newblock Accessed September 22, 2026.

\bibitem[Saini et~al.(2026)Saini, Birkbeck, Wang, Adsumilli, and
  Bovik]{cachedsearch}
Shreshth Saini, Neil Birkbeck, Yilin Wang, Balu Adsumilli, and Alan~C. Bovik.
\newblock Cachedsearch: Training-free cached exploration for test-time search
  in video diffusion, 2026.
\newblock URL \url{https://arxiv.org/abs/2607.23159}.

\bibitem[Song et~al.(2026)Song, Pfister, and Song]{vqqa}
Yiwen Song, Tomas Pfister, and Yale Song.
\newblock Vqqa: An agentic approach for video evaluation and quality
  improvement, 2026.
\newblock URL \url{https://arxiv.org/abs/2603.12310}.

\bibitem[Sun et~al.(2024)Sun, Huang, Liu, Wu, Xu, Li, and Liu]{t2vcompbench}
Kaiyue Sun, Kaiyi Huang, Xian Liu, Yue Wu, Zihan Xu, Zhenguo Li, and Xihui Liu.
\newblock {T2V-CompBench}: A comprehensive benchmark for compositional
  text-to-video generation, 2024.
\newblock URL \url{https://arxiv.org/abs/2407.14505}.

\bibitem[Wan et~al.(2025)Wan, Wang, Ai, Wen, Mao, Xie, Chen, Yu, Zhao, Yang,
  Zeng, Wang, Zhang, Zhou, Wang, Chen, Zhu, Zhao, Yan, Huang, Feng, Zhang, Li,
  Wu, Chu, Feng, Zhang, Sun, Fang, Wang, Gui, Weng, Shen, Lin, Wang, Wang,
  Zhou, Wang, Shen, Yu, Shi, Huang, Xu, Kou, Lv, Li, Liu, Wang, Zhang, Huang,
  Li, Wu, Liu, Pan, Zheng, Hong, Shi, Feng, Jiang, Han, Wu, and Liu]{wan}
Team Wan, Ang Wang, Baole Ai, Bin Wen, Chaojie Mao, Chen-Wei Xie, Di~Chen,
  Feiwu Yu, Haiming Zhao, Jianxiao Yang, Jianyuan Zeng, Jiayu Wang, Jingfeng
  Zhang, Jingren Zhou, Jinkai Wang, Jixuan Chen, Kai Zhu, Kang Zhao, Keyu Yan,
  Lianghua Huang, Mengyang Feng, Ningyi Zhang, Pandeng Li, Pingyu Wu, Ruihang
  Chu, Ruili Feng, Shiwei Zhang, Siyang Sun, Tao Fang, Tianxing Wang, Tianyi
  Gui, Tingyu Weng, Tong Shen, Wei Lin, Wei Wang, Wei Wang, Wenmeng Zhou, Wente
  Wang, Wenting Shen, Wenyuan Yu, Xianzhong Shi, Xiaoming Huang, Xin Xu, Yan
  Kou, Yangyu Lv, Yifei Li, Yijing Liu, Yiming Wang, Yingya Zhang, Yitong
  Huang, Yong Li, You Wu, Yu~Liu, Yulin Pan, Yun Zheng, Yuntao Hong, Yupeng
  Shi, Yutong Feng, Zeyinzi Jiang, Zhen Han, Zhi-Fan Wu, and Ziyu Liu.
\newblock {Wan}: Open and advanced large-scale video generative models.
\newblock \emph{arXiv preprint arXiv:2503.20314}, 2025.
\newblock URL \url{https://arxiv.org/abs/2503.20314}.

\bibitem[Wang et~al.(2025)Wang, Zhou, Yuan, Woo, Gong, Han, Yang, Zhang, Zhu,
  Liu, and Zhao]{consistencysolver}
Fu-Yun Wang, Hao Zhou, Liangzhe Yuan, Sanghyun Woo, Boqing Gong, Bohyung Han,
  Ming-Hsuan Yang, Han Zhang, Yukun Zhu, Ting Liu, and Long Zhao.
\newblock Image diffusion preview with consistency solver, 2025.
\newblock URL \url{https://arxiv.org/abs/2512.13592}.

\bibitem[Yuan et~al.(2026)Yuan, Yin, Li, Huang, Yang, and Yuan]{helios}
Shenghai Yuan, Yuanyang Yin, Zongjian Li, Xinwei Huang, Xiao Yang, and Li~Yuan.
\newblock {Helios: Real Real-Time Long Video Generation Model}, 2026.
\newblock URL \url{https://arxiv.org/abs/2603.04379}.

\bibitem[Zhao et~al.(2023)Zhao, Bai, Rao, Zhou, and Lu]{unipc}
Wenliang Zhao, Lujia Bai, Yongming Rao, Jie Zhou, and Jiwen Lu.
\newblock {UniPC}: A unified predictor-corrector framework for fast sampling of
  diffusion models, 2023.
\newblock URL \url{https://arxiv.org/abs/2302.04867}.

\bibitem[Zhao et~al.(2026)Zhao, Liu, Cao, Gong, Zhang, Song, Deng, and
  Patras]{latsearch}
Zengqun Zhao, Ziquan Liu, Yu~Cao, Shaogang Gong, Zhensong Zhang, Jifei Song,
  Jiankang Deng, and Ioannis Patras.
\newblock Latsearch: Latent reward-guided search for faster inference-time
  scaling in video diffusion, 2026.
\newblock URL \url{https://arxiv.org/abs/2603.14526}.

\end{thebibliography}
